\documentclass[11pt,a4paper]{article}
\usepackage[hyperref]{acl2020}
\usepackage{times}
\usepackage{latexsym}

\usepackage{microtype}

\usepackage{multirow}
\usepackage{graphicx}
\usepackage{amsmath}
\usepackage{enumitem}
\aclfinalcopy 

\title{Selecting Backtranslated Data from Multiple Sources for Improved Neural Machine Translation}
\author{Xabier Soto,\textsuperscript{1} Dimitar Shterionov,\textsuperscript{2} Alberto Poncelas,\textsuperscript{2} and Andy Way\textsuperscript{2} \\
  \textsuperscript{1}Ixa NLP Group, HiTZ Center,  University of the Basque Country (UPV/EHU)\\
  \textsuperscript{2}ADAPT Centre, School of Computing, Dublin City University\\
  \textsuperscript{1}\texttt{xabier.soto@ehu.eus} \\
  \textsuperscript{2}\texttt{\{firstname.lastname\}@adaptcentre.ie} \\}

\date{}

\begin{document}
\maketitle
\begin{abstract}
Machine translation (MT) has benefited from using synthetic training data originating from translating monolingual corpora, a technique known as backtranslation.
Combining backtranslated data from different sources has led to better results than when using such data in isolation. In this work we analyse the impact that data translated with rule-based, phrase-based statistical and neural MT systems has on new MT systems. We use a real-world low-resource use-case (Basque-to-Spanish in the clinical domain) as well as a high-resource language pair (German-to-English) to test different scenarios with backtranslation and employ data selection to optimise the synthetic corpora. We exploit different data selection strategies in order to reduce the amount of data used, while at the same time maintaining high-quality MT systems. We further tune the data selection method by taking into account the quality of the MT systems used for backtranslation and lexical diversity of the resulting corpora. Our experiments show that incorporating backtranslated data from different sources can be beneficial, and that availing of data selection can yield improved performance.

\end{abstract}

\section{Introduction}\label{sec:intro}
The use of supplementary backtranslated text has led to improved results in several tasks such as automatic post-editing \cite{JunczysDowmunt2016log,Hokamp2017ensembling}, machine translation (MT)~\cite{sennrich-2016-improving,poncelas2018investigating}, and quality estimation \cite{Yankovskaya2019quality}. Backtranslated text is a translation of a monolingual corpus in the target language (L2) into the source language (L1) via an already existing MT system, so that the aligned monolingual corpus and its translation can form an L1--L2 parallel corpus. This corpus of synthetic parallel data can then be used for training, typically alongside authentic human-translated data. For MT, backtranslation 
has become a standard approach to improving the performance of systems when additional monolingual data in the target language is available.

While~\citet{sennrich-2016-improving} show that any form of source-side data (even using dummy tokens on the source side) can improve MT performance, both the quality and quantity of the backtranslated data play a significant role in practice. Accordingly, the choice of systems to be used for backtranslation is crucial. In~\citet{poncelas-etal-2019}, different combinations of backtranslated data originating from phrase-based statistical MT (PB-SMT) and neural MT (NMT) were shown to have different impacts on the quality of MT systems. 

In this work we conduct a systematic study of the effects of backtranslated data from different sources, as well as how to optimally select subsets of this data taking into account the loss in quality and lexical richness when data is translated with different MT systems. 
That is, we aim to (i) provide a systematic analysis of backtranslated data from different sources; and (ii) to exploit a reduction in the amount of training data while maintaining high translation quality. 
To achieve these objectives we analyse backtranslated data from several MT systems and investigate multiple approaches to data selection for backtranslated data based on the Feature Decay Algorithms (FDA:~\citet{bicici-2015,FDANMT}) method. We exploit different ways of ranking the data and extracting parallel sentences; we also interleave quality evaluation and lexical diversity/richness information into the ranking process. While our empirical evaluation shows different results for the tested language pairs, this is the first work in this direction and lays a firm foundation for future research.


Nowadays, NMT~\citep{kalchbrenner-blunsom-2013-recurrent-cont,sutskever-etal-2014,bahdanau-2014}, and in particular Transformer~\citep{vaswani-etal-2017} achieves state-of-the-art results for many domains and language pairs. However, NMT requires a lot more data than other paradigms~\citep{Koehn2017-six-challenges}, which makes it harder to adapt to low-resource scenarios~\citep{sennrich-zhang-2019-revisiting-low}. Using synthetic parallel data via backtranslation has been helpful in some low-resource use-cases \cite{Meghan}. 
For extreme cases with no bilingual parallel corpora, unsupervised MT can obtain reasonable results \cite{artetxe-et-al-2019-effective,lample-2019}. However, its application to real low-resource scenarios is still a matter of study \citep{marchisio-2020}. In this work we are motivated by a real-world low-resource use-case, namely the translation of clinical texts from Basque to Spanish (EU-ES). Basque is a minority language, so most of the Electronic Health Records (EHR) are written in Spanish so that any doctor from the Basque public health service can understand them. The development of a system for translating clinical texts from Basque to Spanish could allow Basque-speaking doctors to write EHRs in Basque, thus contributing to the normalisation of the language in specialised areas. 

We conduct our analysis in the scope of the EU-ES translation of EHR use-case, as well as on a language pair and a data set that have been well studied in the literature -- German to English (DE-EN) data used in the WMT Biomedical Translation Shared Task \citep{bawden-et-al-2019-findings}. 
As the EU-ES medical data cannot be made publicly available due to privacy regulations, using the DE-EN data is a way to allow for the replicability of our work. 

\section{Related Work}\label{sec:background}
One of the first papers comparing the performance of different systems for backtranslation was~\citet{burlot-yvon-2018-using-monolingual}. The authors compared SMT and NMT systems, obtaining similar results. Closer to our work,~\citet{soto-et-al-2019-leveraging} also try RBMT, PB-SMT and NMT systems for backtranslating EHRs from Spanish into Basque. However, both papers are limited to comparing the performance of systems trained with backtranslated data originating from a single source, without examining whether a combination might be more effective.

More recently~\citet{poncelas-etal-2019} combined the outputs of PB-SMT and NMT systems used for backtranslation, showing that the combination of synthetic data originating from different sources was useful in improving translation performance. In this work we extend these ideas by combining backtranslated data from RBMT, PB-SMT, NMT (LSTM) and NMT (Transformer); in addition, we use FDA to select sentences translated by different systems and analyse the impact of data selection of backtranslated data on the overall translation performance. Regarding the use of data-selection techniques in conjunction with synthetic data, \citet{poncelas2019selecting} fine-tune NMT models with sentences selected from a backtranslated set, and \citet{chinea2017adapting} select monolingual source-side sentences to generate synthetic target strings to improve the translation model.


While the most common approach to assessing the translation capabilities of a MT system is via evaluation scores such as BLEU~\citep{papineni-et-al-2002-bleu}, TER~\citep{snover-etal-2006}, chrF~\citep{popovic-2015chrf}, and METEOR~\citep{banerjee-lavie-2005}, recently research has begun to address another side of quality of translated text, namely lexical richness and diversity. In a recent paper, ~\citet{vanmassenhove-2019-lost} study the loss of lexical diversity and richness of the same corpora translated with PB-SMT and NMT systems. \citet{vanmassenhove-2019-lost} investigate the problem for seen (during MT training) and unseen text using MT systems trained on the Europarl corpus~\citep{Koehn2005-Europarl}, with original (human-produced and translated) text as well as in a round-trip-translation setting.\footnote{In their experiments, \citet{vanmassenhove-2019-lost} backtranslate the training data via an MT system trained on the same data, then train yet another system with this data and analyse its performance. They assess how errors propagate through repeated translation, thereby investigating the extent of inherent algorithm bias in MT models.} 
In this work we calculate the same lexical diversity metrics as~\citet{vanmassenhove-2019-lost}, and further use those metrics to improve the data selection process applied to backtranslated data.

\section{Data Selection for Backtranslation from Multiple Sources}\label{sec:ds_for_bt}

FDA~\citep{bicici-2015,FDANMT} is a data selection technique that retrieves sentences from a corpus based on the number of {\em n}-grams overlapping with those present in an in-domain data set referred to as $S_{seed}$. 
FDA scores each candidate sentence $s$ according to: (i) the number of {\em n}-grams that are shared with the seed $S_{seed}$; and (ii) the {\em n}-grams already present in a set L of selected sentences, as defined in \eqref{eq:fda_sentencescore}:

\begin{small}
\begin{equation}[t]\label{eq:fda_sentencescore}
score(s,S_{seed},L)=\frac{\sum_{ngr \in \{s \bigcap  S_{seed} \} } 0.5^{C_L(ngr)} }
{\text{length(s)}}
\end{equation}
\end{small}

\noindent where $length(s)$ is the number of words in the sentence $s$ and $C_L(ngr)$ is the number of occurrences of the {\em n}-gram $ngr$ in $L$. The score is then used to rank sentences, with the one with the highest score being selected and added to $L$. This process is repeated iteratively. To avoid selecting sentences containing the same {\em n}-grams, $score(s,S_{seed},L)$ applies a penalty to the {\em n}-grams (up to order three in the default configuration) proportional to the occurrences that have been already selected. In~\eqref{eq:fda_sentencescore}, the term $0.5^{C_L(ngr)}$ is used as the penalty.

In the context of MT, FDA has been shown to obtain better results than other methods for data selection~\citep{silva-etal-2018}. Acordingly, in this work we too focus on FDA, although our rescoring idea is more general and can be applied to other selection methods based on {\em n}-gram overlap.

Related work on quality and lexical diversity and richness of MT demonstrates that (i) regardless of the overall performance of an MT system (as measured by both automatic and human evaluation), in general machine-translated text is error-prone and cannot reach human quality \cite{toral-2018-attaining}); and (ii) machine-translated text lacks the lexical richness and diversity of human-translated (or post-edited) text \cite{vanmassenhove-2019-lost}. 

In its operation, FDA compares two types of text -- the seed and the candidate sentences -- without taking into account the quality or the lexical diversity/richness of the candidate text. Our hypothesis is that when selecting data from different sources, FDA cannot account for the differences in quality and lexical diversity/richness of these texts, with the consequence that the selected set ($L$) is sub-optimal. 


We test our hypothesis by assessing the quality and lexical diversity/richness of the backtranslated data with the four different systems as well as with different selected subsets of training data. 

To tackle the problem of sub-optimal FDA-selected datasets, we propose to rescore FDA scores based on quality evaluation and lexical diversity/richness scores.\footnote{We talk about “rescoring” as if we compare equations (1) and (2), the only difference is the rescoring produced by multiplying equation (1) (left part in equation (2)) by the factors dependent on MT quality and lexical diversity (right part in equation (2)).} That is, for each sentence $s_i^{BT}$ from a backtranslated corpus $D_i^{BT}$ originating from the $i^{th}$ MT system, we factor in the quality expressed by the evaluation metrics, $q(D_i^{BT})$ and the lexical diversity/richness expressed by the diversity metrics, $d(D_i^{BT})$ as shown in \eqref{eq:fda_refactoring}:

\begin{small}
\begin{equation}\label{eq:fda_refactoring}
\begin{split}
~& score(s_i^{BT},S_{seed},L)= \\
& \frac{\sum_{ngr \in \{s \bigcap  S_{seed} \} } 0.5^{C_L(ngr)}}
{\text{length(s)}} \cdot \phi(q(D_i^{BT}), d(D_i^{BT})) \\
\end{split}
\end{equation}
\end{small}

\noindent where $\phi$ is a function over quality and lexical diversity metrics producing a non-negative real number. 

We note three considerations with respect to our approach to Equation~\eqref{eq:fda_refactoring}.

\begin{enumerate}[leftmargin=*,nolistsep]
    \item {\bf Sentence-level selection versus document-level quality and lexical diversity/richness evaluation.} The FDA algorithm works on a sentence level, while our approach rescores the FDA scores using document-level metrics. As our goal is to differentiate between the output of different MT systems, we consider metrics that reflect the overall quality of each system. Furthermore, metrics for lexical diversity/richness as type/token ratio (TTR)~\citep{Templin1975certain}, Yule's I~\citep{yule-1944}, and the measure of textual lexical diversity (MTLD)~\citep{Mccarthy2005assessment} are to be calculated on a document-level; the same is valid for automatic evaluation metrics such as BLEU and TER.
    \item {\bf Combined metrics.} We conduct our analysis using the quality metrics BLEU, TER, METEOR and chrF; and TTR, MTLD and Yule's I for lexical diversity/richness. For rescoring we use only BLEU, TER and MTLD as a factor: $\phi = log(BLEU * (100 - TER ) * MTLD)$. We decided on this rescoring formula based on preliminary experiments, as it led to the selection of more sentence pairs originating from models trained with backtranslated data from the system that performs best (for both ES-EU and EN-DE); we chose MTLD based on the findings of~\citet{vanmassenhove-2019-lost} which show this metric to be more suitable for comparative analysis, as well as mitigating issues related to sentence length typical for TTR and Yule's I~\citep{Mccarthy2005assessment}.
    \item {\bf Use of devset as a seed}. Using a development set in MT aims to test whether the performance of the MT system has reached a certain level. In FDA for MT, we use a devset as the seed. In our method we compute BLEU and TER on the devset also used as a seed; MTLD is computed on the backtranslated text, i.e. the synthetic source text.
\end{enumerate}


\section{Language Pairs -- Challenges and Objectives}\label{sec:challenges}
As a challenging low-resource scenario, we chose the translation of clinical texts from Basque to Spanish, for which there is no in-domain bilingual corpora. We make use of available EHRs in Spanish coming from the hospital of Galdakao-Usansolo to create a synthetic parallel corpus via backtranslation. The Galdakao-Usansolo EHR corpus consists of 142,154 documents compiled between 2008 and 2012. After deduplication, we end up with a total of $2,023,811$ sentences.\footnote{Due to privacy requirements, this corpus is not publicly available. Prior to use, it was de-identified by reordering sentences, and only authors who had previously signed a non-disclosure commitment had access to it.}

As a basis for training the MT systems for backtranslation, we use a bilingual out-of-domain corpus of 4.5M sentence pairs: 2.3M sentence pairs from the news domain~\citep{etchegoyhen-2016}, and 2.2M from administrative texts, web-crawling and specialised magazines.

In order to adapt the systems to the clinical domain, we used a bilingual dictionary previously used for automatic clinical term generation in Basque ~\citep{perez-vinaspre-2017}, consisting of 151,111 terms in Basque corresponding to 83,360 unique terms in Spanish.

To evaluate our EU-ES systems, we use EHR templates in Basque written with academic purposes~\citep{etxeberri-2014} together with their manual translations into Spanish produced by a bilingual doctor. These 42 templates correspond to diverse specializations, and were written by doctors of the Donostia Hospital. After deduplication, we obtain 1,648 sentence pairs that are randomly divided into 824 sentence pairs for validation (devset) and 824 for testing.

In order to test the generalisability of our idea, we use a well-researched language pair, German-to-English. 
As our out-of-domain corpus, we used the DE-EN parallel data provided in the WMT 2015 \citep{bojar-EtAl:2015:WMT} news translation task.

The adaptation of systems to the medical domain with backtranslated data is performed using the UFAL data collection.\footnote{\url{https://ufal.mff.cuni.cz/ufal_medical_corpus}} We selected the following subsets: ECDC, EMEA, EMEA\_new\_crawl, MuchMore, PatTR\_Medical and Subtitles. The total amount of sentences was 2,555,138 which after deduplication was reduced to 2,335,892. After filtering misaligned and empty lines,\footnote{We used the clean-corpus-n.pl script provided with the Moses toolkit \citep{koehn-et-al-2007-moses}.} the resulting amount was 2,322,599 sentences. We used the EN monolingual side. For development and test sets we used the Cochrane and NHS 24 subsets from the Himl 2017 set.\footnote{\url{http://www.himl.eu/test-sets}}

Table~\ref{tab:resources} provides the statistics of our corpora.

\begin{table}[h]
  \centering
  \captionsetup{justification=centering}
  {\small \begin{tabular}{c l l l l}
    \hline
     & \textbf{Desc.} & \textbf{Sent.} & \multicolumn{2}{c}{\textbf{Tokens}} \\ 
    &  &  & src  & trg \\\hline
    
    \multirow{4}{*}{\rotatebox[origin=c]{90}{\centering	EU-ES}}
    & out-of-domain & 4.5M & 73M & 102M  \\
    & clinical terms & 151K & 271K & 258K  \\
    & EHRs & 2M & & 33M \\
    & EHR templates & 1.6K & 18.5K & 17.6K  \\ \hline 
    \multirow{4}{*}{\rotatebox[origin=c]{90}{\centering	DE-EN}}
    & out-of-domain & 4.5M & 110M & 116M \\
    & in-domain & 2.3M & & 97M \\
    & devset & 1K & 16K & 15K \\
    & test set & 467 & 10K & 9.7K \\
    \hline
  \end{tabular}}
  \caption{Description and statistics of the used corpora.}
  \label{tab:resources}
\end{table}

\section{Empirical Evaluation}\label{sec:experiments}
Via a set of experiments, we (i) investigate the differences in the backtranslated data originating from the four different MT systems and their impact on the performance of MT systems using this backtranslated data, and (ii) test our hypothesis as well as different approaches to rescoring the data selection algorithm. 



\subsection{Systems Used for Backtranslation}\label{sec:backtranslation_systems}
First, we train PB-SMT, LSTM and Transformer models for the ES-EU and EN-DE (i.e. \textit{reverse}) language directions. Then we backtranslate the monolingual corpus into the target language (EU and DE, respectively) using those systems, as well as a RBMT one. \\
\textbf{RBMT:} We use Apertium~\citep{forcada-etal-2011} for the EN-DE language pair, and Matxin \citep{mayor-2007} for ES-EU, adapted to the clinical domain by the inclusion of the same dictionaries used to train the other systems.\\
\textbf{PB-SMT:} We use Moses with default parameters, using MGIZA for word alignment \cite{Giza}, an ``msd-bidirectional-fe'' lexicalised reordering model and a KenLM \citep{heafield-2011} 5-gram target language model. We tuned the model using Minimum Error Rate Training \cite{MERT} with an n-best list of length 100.\\
\textbf{LSTM:} We use an RNN of 4 layers, with LSTM units of size 512, dropout of 0.2 and a batch-size of 128. We use Adam~\citep{kingma-2014} as the learning optimiser, with a learning rate of 0.0001 and 2,000 warmup steps.\\
\textbf{Transformer:} We train a Transformer model with the hyperparameters recommended by OpenNMT,\footnote{\url{http://opennmt.net/OpenNMT-py/FAQ.html\#how-do-i-use-the-transformer-model} (Accessed on December 9, 2019.)} halving the batch-size so that it could fit in 2 GPUs, and accordingly doubling the value for gradient accumulation.

We train all NMT systems using OpenNMT~\citep{klein-etal-2017} for a maximum of 200,000 steps, and select the model that obtains the highest BLEU score on the devset; note that the final systems trained after applying data selection use early stopping with perplexity not decreasing in 3 consecutive steps as our stopping criterion. Backtranslation is performed with the default hyperparameters, including a beam-width of 5 and a batch-size of 30.

We use Moses scripts to tokenise and truecase all the corpora to be used for statistical or neural systems. For the NMT systems, we apply BPE~\citep{sennrich-etal-2015} on the concatenated bilingual corpora with 90,000 merge operations for EU-ES and 89,500 for DE-EN, using subword-nmt.\footnote{\url{https://github.com/rsennrich/subword-nmt} (Accessed on December 9, 2019.)} 

\subsection{Systems with Data Selected via Backtranslation}\label{sec:ds_anb_bt_systems}
For each language pair we train four Transformer models with the authentic and backtranslated data, as well as a fifth system with all four backtranslated versions concatenated to the authentic data. These we refer to as +\textit{S}$_{bt}$, where \textit{S} is one of RBMT, PB-SMT, LSTM or Transformer and indicates the origin of the backtranslation, and +All$_{bt}$ to refer to the system trained with all backtranslated data.

Next, we use the devset as a seed for the data selection algorithm. Given that FDA does not score sentences that have no {\em n}-gram overlaps with any sentence from the seed, for the ‘EachFromAll’ configuration presented later, which is constrained to select one sentence for each sentence in the monolingual corpus, we randomly select one sentence among those produced by the 4 different systems used for backtranslation, in case none of them overlap with any sentence from the seed. We obtain the FDA scores and use them to order the sentence pairs in descending order. Next, we apply the following different data selection configurations:
\begin{enumerate}[leftmargin=*,nolistsep]
    \item Top from all sentences (referred to as \textit{FromAll} henceforth): concatenate the data backtranslated with all the systems and select the top ranking 2M (for EU-ES) or 2.3M (for DE-EN) sentence pairs with the possibility of selecting the same target sentence more than once, i.e. translated by different systems.
    \item Top for each (target) sentence (henceforth, \textit{EachFromAll}): concatenate the data backtranslated with all the systems and select the optimal sentence pairs avoiding the selection of the same target sentence more than once. That is, each selected target sentence will have only one associated source sentence originating from one specific system.
    \item Top for each (target) sentence x4 (henceforth, \textit{EachFromAll x4}): same as EachFromAll, but repeating the selected backtranslated data four times (only for EU-ES).
    \item Top for each (target) sentence \textbf{rescored} (henceforth, \textit{EachFromAll RS}): use MT evaluation and lexical diversity metrics to rescore the FDA ranks and perform an EachFromAll selection.
\end{enumerate} 

We selected the Transformer architecture as the basis of our backtranslation models because (i) it has obtained the best performance for many use-cases and language pairs which we also aim at, and (ii) it has been shown that Transformer's performance is strongly impacted by the quantity of data, which can act as an indicator as to whether our improvements originate from the quantity or the quality of the data. That is why we compare EachFromAll systems to systems trained with all backtranslated data (i.e. all 8M sentence pairs), to verify that it is not only the amount of data that impacts performance. 

\section{Results and Analysis}\label{sec:results-and-analysis}
\subsection{MT Evaluation}\label{sec:mt-evaluation}
We use the automatic evaluation metrics BLEU, TER, METEOR and chrF (in its chrF3 variant) to assess the translation quality of our systems. In Table~\ref{tab:backtranslation-results} we show the scores on the test set of the \textit{reverse} systems used for backtranslation (the best are marked in bold). For EU-ES, since we only use clinical terms as in-domain training data, the results are poor overall. However, we observe that Transformer obtains the best results according to all metrics for both EU-ES and DE-EN. 
Table~\ref{tbl:baseline-systems} shows the results of our baseline (\textit{forward}) systems. It shows that Transformer systems perform best for both language pairs. 
Evaluation scores for the systems trained on authentic and backtranslated data, and for the systems trained after data selection for EU-ES and DE-EN, are shown in Table~\ref{tbl:back_dataselection_results}. 

\begin{table}[th]
\centering \setlength\tabcolsep{3.3pt}
{\small \begin{tabular}{clrrrr}
  & & BLEU$\uparrow$    & TER$\downarrow$    & METEOR$\uparrow$  & CHRF3$\uparrow$ \\ \hline
\multirow{4}{*}{\rotatebox[origin=c]{90}{ES-EU}} & RBMT & 11.37   & 75.52  & 19.80   & 41.35  \\
& PB-SMT & 9.38    & 70.70  & 25.36   & 44.07 \\
& LSTM & 7.01    & 72.29  & 20.46   & 33.94 	\\
& Transformer & \textbf{12.21}   & \textbf{66.53}  & \textbf{26.96}   & \textbf{44.42}  \\ \hline
\multirow{4}{*}{\rotatebox[origin=c]{90}{EN-DE}} & RBMT &   8.21    &   72.26    &    25.70      &    41.40     \\
& PB-SMT &    14.85	&	74.00	&	35.62	&	48.92	\\
& LSTM &     24.65	&	54.60	&	43.30	&	53.51	\\
& Transformer & \textbf{32.24}	&	\textbf{46.83}	&	\textbf{50.25}	&	\textbf{60.29}\\ \hline
\end{tabular}}
\caption{Scores of \textit{reverse} systems for backtranslation. }
\label{tab:backtranslation-results}
\end{table}

\begin{table}[th]
\centering \setlength\tabcolsep{3.3pt}
{\small \begin{tabular}{clrrrr}
& & BLEU$\uparrow$    & TER$\downarrow$    & METEOR$\uparrow$  & CHRF3$\uparrow$ \\ \hline
\multirow{2}{*}{\rotatebox[origin=c]{90}{{\scriptsize EU-ES}}} & LSTM  & 10.84 & 85.00 & 32.79 & 41.36\\
& Transformer  & \textbf{19.64} & \textbf{69.11} & \textbf{43.84} & \textbf{53.03}\\\hline
\multirow{2}{*}{\rotatebox[origin=c]{90}{{\scriptsize DE-EN}}} &  LSTM  & 28.15 & 51.95 & 32.19 & 55.40\\
&  Transformer  & \textbf{38.27} & \textbf{42.87} & \textbf{37.02} & \textbf{62.37}\\\hline
\end{tabular}}
\caption{Scores of baseline systems. }
\label{tbl:baseline-systems}
\end{table}

\begin{table}[bh]
    \centering
    {\small \setlength\tabcolsep{1.8pt}
    \begin{tabular}{cclrrrr}
  & & & BLEU$\uparrow$    & TER$\downarrow$    & MET.$\uparrow$  & CHRF3$\uparrow$ \\ \hline
\multirow{10}{*}{\rotatebox[origin=c]{90}{EU-ES}} & & +RBMT$_{bt}$  & 23.27 & 62.67 & 48.02 & 56.51\\
& Auth. &  +PB-SMT$_{bt}$  & 22.51 & 64.57 & 45.97 & 54.53\\
& + & +LSTM$_{bt}$  & 24.74 & 63.55 & 47.58 & 55.59\\
& BT. &  +Transformer$_{bt}$  & 25.70 & 60.29 & 48.53 & 57.08\\
& & +All$_{bt}$  & \textbf{26.18} & \textbf{59.10} & \textbf{49.19} & \textbf{57.31}\\\cline{2-7}
& Auth. &  FromAll  & \textbf{\textit{25.93}} & 59.76 & 48.66 & 56.69\\
& BT. &  EachFromAll  & 25.85 & \textbf{\textit{58.92}} & \textbf{\textit{48.83}} & \textbf{\textit{57.17}}\\
& + &  EachFromAll x4 & 24.59 & 61.15 & 48.10 & 56.19\\
& DS & EachFromAll RS & 25.77 & 59.86 & 48.59 & 56.92\\\hline\hline

\multirow{8}{*}{\rotatebox[origin=c]{90}{DE-EN}} & &  +RBMT$_{bt}$  & 39.02 & 42.27 & 37.32 & 62.72\\
& Auth. &  +PB-SMT$_{bt}$  & 42.32 & 39.21 & 39.37 & 65.91\\
& + &  +LSTM$_{bt}$  & 40.97 & 39.75 & 38.45 & 64.81\\
& BT  &  +Transformer$_{bt}$  & \textbf{42.75} & 38.73 & 39.35 & \textbf{66.05}\\
& &  +All$_{bt}$  & 42.69 & \textbf{38.45} & \textbf{39.65} & 65.99\\\cline{2-7}
& Auth. &  FromAll  & 43.66 & \textbf{\textit{37.71}} & \textbf{\textit{40.10}} & 67.01\\
& + BT &  EachFromAll  & 43.45 & 38.24 & 39.81 & 66.44\\
& + DS & EachFromAll RS &	\textbf{\textit{43.98}}	&	37.79	&	39.91	&	\textbf{\textit{67.10}}	\\\hline
\end{tabular}}
    \caption{Scores for systems trained on authentic (Auth.) and backtranslated (BT) data, and after data selection (DS). MET. abbreviates METEOR.}
    \label{tbl:back_dataselection_results}
\end{table}

We observe from Table~\ref{tbl:back_dataselection_results} that for both language pairs the inclusion of backtranslated data clearly improves the results of the baseline systems. For EU-ES the ordering of the systems from best to worse is Transformer $>$ RBMT $>$ LSTM $>$ PB-SMT for all metrics except BLEU, where the order is Transformer $>$ LSTM $>$ RBMT $>$ PB-SMT. The EU-ES system trained on (authentic data and) data translated by all systems (+All$_{bt}$), thus using 4 times more backtranslated data than the rest, obtains the best results; however, the observed improvements are not as high as those for the other systems, e.g. the best (+Transformer$_{bt}$) has a 0.96 BLEU point improvement over the second best (+LSTM$_{bt}$), while the +All$_{bt}$ system is only $0.48$ BLEU points better than +Transformer$_{bt}$. This tendency is the same for the other metrics too. For the DE-EN use-case the score differences between the best systems (+Transformer$_{bt}$ or +PB-SMT$_{bt}$ depending on the metric) and +All$_{bt}$ are even smaller, with BLEU and chrF3 favouring the former, and TER and METEOR the latter. 


For EU-ES, all systems trained with 2M sentence pairs selected from the backtranslated data according to the basic DS methods and the newly proposed method with rescoring obtain better results than any system trained with backtranslated data originating from a \textit{single} system. Furthermore, according to all metrics except BLEU, the EachFromAll system outperforms FromAll. Compared to the system including the data translated by all systems (+All$_{bt}$), EachFromAll is better only in terms of TER. These results show that either the quantity of data leads to differences in performance (comparing the best system after data selection, i.e. EachFromAll, to +All$_{bt}$), or that the data selection method fails to retrieve those sentence pairs that would lead to better performance. In order to test these two assumptions, we first train a system with the EachFromAll data repeated 4 times resulting in the same number of sentence pairs as in the +All$_{bt}$ case. According to the resulting evaluation scores, this system is worse than +All$_{bt}$, but also worse than any of the basic data selection configurations. This indicates that the diversity (among the source sentences) gained by using 4 different systems for backtranslation is more important than the quantity of the data in terms of automatic scores. While for EU-ES the EachFromAll selection configuration achieves the best results, for DE-EN the FromAll configuration leads to better scores. Furthermore, this configuration outperforms the system with all backtranslated data (+All$_{bt}$). 

Next, we train a system with data selected from the backtranslated data after the original FDA scores have been rescored using the quality and lexical diversity/richness scores. These systems are shown in Table~\ref{tbl:back_dataselection_results} with the suffix RS (i.e. ReScored). While for EU-ES this system does not outperform the rest, in the DE-EN case we observe that it does. With the exception of the TER and METEOR scores, the EachFromAll RS for the DE-EN language pair is the best system. These experiments show different outcomes for each language pair and thus disagree with respect to our hypothesis of rescoring the data selection scores being beneficial for MT. Accordingly, more experiments are needed to specify how to perform this rescoring, as well as in which settings our rescoring proposal is beneficial. 
Further analysis and a discussion on lexical diversity/richness, data selection and sentence length follow in the rest of this section.


\subsection{Lexical Diversity/Richness}\label{sec:lexical-diversity}
We analyse the lexical diversity/richness of the corpora of both language pairs based on the Yule's I, MTLD and TTR metrics. We calculate these scores for the corpora resulting from backtranslation by the different systems (BT), for the corpora resulting from applying the basic data selection approaches (DS), and the development and test sets used for evaluation (EV). We show these scores in Table~\ref{tab:lexical-diversity-results-eu-es} and Table~\ref{tab:lexical-diversity-results-de-en} for EU-ES and DE-EN, respectively.

Regarding the different systems used for backtranslation, we observe that for EU-ES the sentences translated by the RBMT system are much more diverse than the rest according to all metrics, while Transformer obtains the highest scores among the other three. For the DE-EN corpora, this is not the case, and the data from the Transformer system is more diverse according to Yule's I and TTR, but not according to MTLD. 

\begin{table}[t]
\hspace{-0.5cm}
\begin{center}
\captionsetup{justification=centering}
\setlength\tabcolsep{2pt}
{\small \begin{tabular}{ c l c c c c c c }\hline
Type & \multicolumn{1}{c}{Corpus} & \multicolumn{2}{c}{Yule's I*100} & \multicolumn{2}{c}{MTLD} & \multicolumn{2}{c}{TTR * 100} \\ 
&  & EU & ES & EU & ES & EU & ES \\ \hline
\multicolumn{1}{c}{\multirow{4}{*}{BT}} & RBMT$_{bt}$ & 74.3 & \multirow{4}{*}{0.91} & 15.33 & \multirow{4}{*}{14.06} & 3.70 & \multirow{4}{*}{1.01} \\ 
\multicolumn{1}{c}{} & PB-SMT$_{bt}$ & 0.40    &  & 13.76 & & 1.01 &  \\ 
\multicolumn{1}{c}{} & LSTM$_{bt}$ & 3.23   &  & 13.20 & & 2.77 &  \\ 
\multicolumn{1}{c}{} & Trans.$_{bt}$     & 8.19   &   & 13.79 & \\ \hline
\multicolumn{1}{c}{\multirow{3}{*}{DS}} & FA  & 2.81 & 0.16 & 13.73 & 13.91 & 2.26 & 0.42 \\ 
\multicolumn{1}{c}{} & EFA & 5.78 & 0.91 & 13.88 & 14.03 & 3.08 & 1.01  \\
\multicolumn{1}{c}{} & EFA RS & 9.54 & 0.91 & 13.84 & 14.03 & 3.67 & 1.01 \\ \hline 
\multicolumn{1}{c}{\multirow{2}{*}{EV}} & Dev. & 626 & 456 & 13.72 & 13.92 & 32.90 & 27.50  \\ 
\multicolumn{1}{c}{} & Test & 663 & 491 & 13.63 & 13.75 & 32.80 & 27.50 \\ \hline
\end{tabular}}
\end{center}
\caption{Lexical diversity scores of the backtranslation (BT), data selection (DS) and evaluation (EV) corpora for the ES-EU and EU-ES systems. Trans. = Transformer, FA = ForAll, EFA = EachFromAll, EFA RS = EachFromAll Rescored.}
\label{tab:lexical-diversity-results-eu-es}
\end{table} 
 
\begin{table}[t]
\hspace{-0.5cm}
\begin{center}
\captionsetup{justification=centering}
\setlength\tabcolsep{2pt}
{\small \begin{tabular}{ c l c c c c c c }\hline
Type & \multicolumn{1}{c}{Corpus} & \multicolumn{2}{c}{Yule's I*100} & \multicolumn{2}{c}{MTLD} & \multicolumn{2}{c}{TTR * 100} \\ 
&  &  DE & EN	&	DE & EN	&	DE & EN \\ \hline
\multicolumn{1}{c}{\multirow{4}{*}{BT}} & RBMT$_{bt}$ & 4.55 &	\multirow{4}{*}{2.68} & 48.50 & \multirow{4}{*}{37.50}	& 1.64 & \multirow{4}{*}{1.56} \\ 
\multicolumn{1}{c}{} & PB-SMT$_{bt}$ & 0.66	& &	74.90 &	&	0.80 &  \\ 
\multicolumn{1}{c}{} & LSTM$_{bt}$ & 2.31	& &	40.00 &	&	1.90 &  \\ 
\multicolumn{1}{c}{} & Trans.$_{bt}$     & 5.62	& &	53.70	& &	2.61 &  \\ \hline
\multicolumn{1}{c}{\multirow{3}{*}{DS}} & FA  & 2.49 & 0.11	& 107.00 &	50 &	1.44 & 0.36  \\ 
\multicolumn{1}{c}{} & EFA & 3.96	& 0.39 &	103.00	&	46.00 & 1.83 & 0.69   \\
\multicolumn{1}{c}{}	& EFA RS & 5.39 & 0.39 & 105.00	& 45.60 & 2.56 & 0.69	\\ \hline 
\multicolumn{1}{c}{\multirow{2}{*}{EV}} & Dev & 386 & 282 &	108.15 & 61.06 & 20.00 & 15.59  \\ 
\multicolumn{1}{c}{} & Test & 528 & 301 & 117.90 & 59.63 & 23.83 & 18.11 \\ \hline
\end{tabular}}
\end{center}
\caption{Lexical diversity scores of the backtranslation (BT), data selection (DS) and evaluation (EV) corpora for the EN-DE and DE-EN systems. Trans. = Transformer, FA = ForAll, EFA = EachFromAll, EFA RS = EachFromAll Rescored.}
\label{tab:lexical-diversity-results-de-en}
\end{table}

We note that Yule’s I and TTR depend on the amount of sentences in the assessed corpora. As such, we can see that for the development and test sets the scores are quite a bit higher than the rest. Accordingly, comparisons should be only be conducted for corpora with the same number of sentences.

Following the analysis and discussion in~\citet{vanmassenhove-2019-lost}, we decided to use MTLD as the lexical diversity metric for our rescoring data selection approach, as defined in Section~\ref{sec:ds_for_bt}. 


\subsection{Systems Selected by Data Selection}\label{sec:selected-systems}
\begin{figure}[ht]
    \centering
\includegraphics[width=0.43\textwidth]{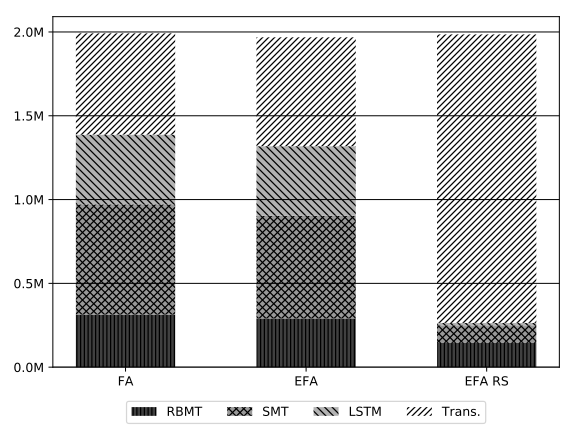}
    \caption{Amount of sentences selected from each system by the data selection approaches for EU-ES. FA = FromAll, EFA = EachFromAll, EFA RS = EachFromAll Rescored.}
    \label{fig:eu-es_selected}
\end{figure}

\begin{figure}[ht]
    \centering
\includegraphics[width=0.43\textwidth]{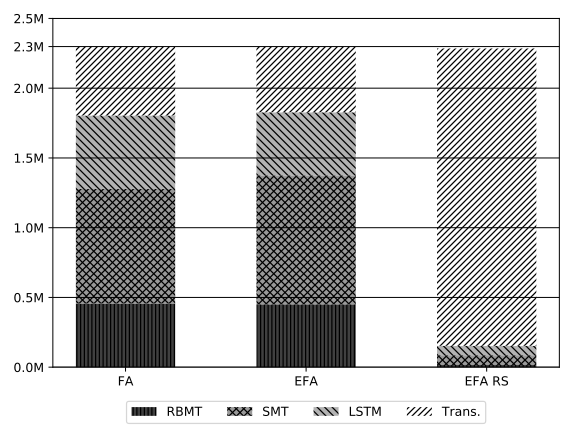}
    \caption{Amount of sentences selected from each system by the data selection approaches for EN-DE. FA = FromAll, EFA = EachFromAll, EFA RS = EachFromAll Rescored.}
    \label{fig:en-de_selected}
\end{figure}

We first analyse how the basic data selection methods choose different numbers of sentences from each system used for backtranslation, and then we compare them with the rescoring method. Figures~\ref{fig:eu-es_selected} and~\ref{fig:en-de_selected} show the portion of selected sentences per backtranslation system that form the training sets for the systems listed in Table~\ref{tbl:back_dataselection_results}. 

For EU-ES, we observe that the EachFromAll configuration (the one with the highest scores according to the evaluation metrics in Table~\ref{tbl:back_dataselection_results}) selects more sentences from Transformer (649,312) in contrast to the ForAll approach that prefers PB-SMT (657,543). For DE-EN, FromAll and EachFromAll tend to select a higher number of sentences backtranslated by the PB-SMT model (820,765 and 924,694, respectively). However, for both language pairs, both ForAll and EachFromAll distributions are very similar as can be seen in Figures~\ref{fig:eu-es_selected} and~\ref{fig:en-de_selected}. Given that the DE-EN system trained with backtranslated data from PB-SMT (+PB-SMT$_{bt}$) obtains the worst results while the one from Transformer (+Transformer$_{bt}$) performs the best, we correlate the two measurements and hypothesise that a distribution where more sentences originating from Transformer are selected would yield better results. Our $\phi$ rescoring (cf. Equation~\eqref{eq:fda_refactoring}) shifts the preferred selection system to Transformer. For EU-ES, the EachFromAll Rescored selects 1,720,736 out of the total of 1,985,227 sentences (about 87\%); for DE-EN, it selects 2,131,227 out of the total of 2,284,800 sentences (93\%).

\begin{figure}[t]
\centering
\includegraphics[width=0.43\textwidth]{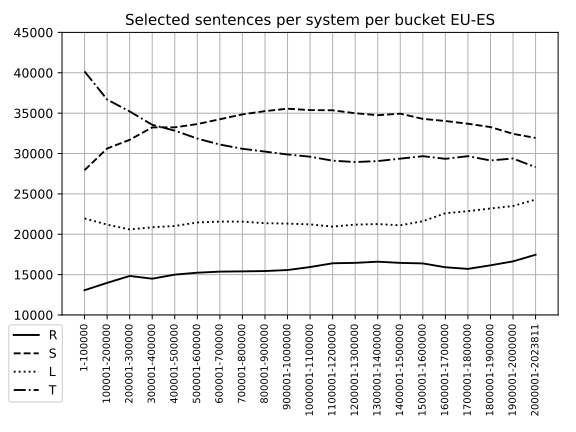}
\caption{Number of sentences selected from each system by the FromAll data selection approach for EU-ES language pair in subsequent bins of 100,000 sentences (extrapolated for the last bin).}
\label{fig:selected-systems-figure}
\end{figure}

\begin{figure}[t]
\centering
\includegraphics[width=0.43\textwidth]{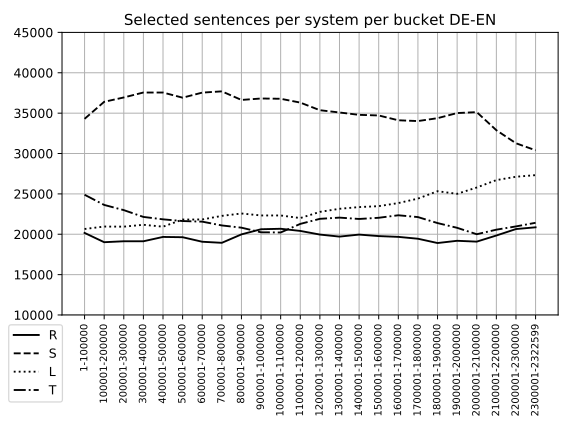}
\caption{Number of sentences selected from each system by the FromAll data selection approach for DE-EN language pair in subsequent bins of 100,000 sentences (extrapolated for the last bin).}
\label{fig:selected-systems-figure_de_en}
\end{figure}

For a more in-depth view of the distribution of selected sentence pairs per backtranslation system, we present the amount of selected sentences per system in bins of 100,000 for the FromAll systems. We show the results for EU-ES in Figure~\ref{fig:selected-systems-figure} and for DE-EN in Figure~\ref{fig:selected-systems-figure_de_en}. 
For EU-ES, we observe that Transformer is the most selected system for the first bins, but the number of sentences sharply decreases until the middle of the corpus and then stabilises. In contrast, the number of sentences originating from PB-SMT increases in the first half and slowly decreases afterwards. The number of sentences from RBMT and LSTM seams more stable, with a slight tendency to increase, peaking in the last bins. For DE-EN, we observe that PB-SMT is always the preferred system, but with a decreasing tendency; and the number of sentences originating from LSTM increases towards the last bins.

\subsection{Sentence Length}\label{sec:sentence-length}
We also analyse how the average sentence length varies during the data selection process in the FromAll configuration, as we did in Section~\ref{sec:selected-systems} when analysing the selected systems. 

Table~\ref{tab:sentence-length-results-eu-es} shows the average sentence lengths of the EU-ES and DE-EN data from the different reverse systems (BT), of the corpora resulting after data selection (DS) and of the test and the development sets (EV). We note that the sentences translated by PB-SMT are longer than those translated by any other system for both language pairs. Correlating these results with those presented in Table~\ref{tbl:back_dataselection_results} and in Figures~\ref{fig:selected-systems-figure} and~\ref{fig:selected-systems-figure_de_en}, we can assert that in FDA the length penalty has a weaker effect than {\em n}-gram overlap and as such FDA has a preference towards {\em n}-gram MT paradigms, i.e. PB-SMT. However, data selection that results in more Transformer sentences would appear to be a better option.



\begin{table}[h]
\centering
\setlength\tabcolsep{5pt}
{\small \begin{tabular}{c l r r r r} 
\hline
Type & \multicolumn{1}{c}{Corpus} & \multicolumn{1}{c}{EU}  & \multicolumn{1}{c}{ES}    & \multicolumn{1}{c}{DE}    & \multicolumn{1}{c}{EN}    \\ \hline
\multirow{4}{*}{BT}	&	RBMT$_{bt}$	&	10.56	&	16.16	&	33.64	&	34.30	\\ 	
&	PB-SMT$_{bt}$	&	16.09	&	16.16	&	39.04	&	34.30	\\ 	
&	LSTM$_{bt}$ 	&	12.53	&	16.16	&	29.55	&	34.30	\\ 	
&	Transformer$_{bt}$	&	12.62	&	16.16	&	23.37	&	34.30	\\	\hline
\multirow{2}{*}{DS}	&	FromAll	&	17.60	&	21.21	&	41.61	&	51.84	\\	
&	EachFromAll	&	13.67	&	16.16	&	32.94	&	34.30	\\	\hline
\multirow{2}{*}{EV}	&	Dev. 	&	10.85	&	10.34	&	15.09	&	14.34	\\	
&	Test	&	11.64	&	11.04	&	21.27	&	20.79	\\	\hline
\end{tabular}}
\caption{Average sentence length of the backtranslation (BT), data selection (DS) and evaluation sets (EV).}
\label{tab:sentence-length-results-eu-es}
\end{table}



\section{Conclusions and Future Work}\label{sec:conclusions}

We evaluated several approaches to data selection over the data backtranslated by RBMT, PB-SMT, LSTM and Transformer systems for two language pairs (EU-ES and DE-EN) from the clinical/biomedical domain.
The former is a low-resource language pair, and the latter a well researched, high-resource language pair. 
Furthermore, in terms of the two target languages, English is a morphologically less rich language than Spanish, which creates a different setting again in which to evaluate our methodology. We use these two different use-cases to better understand both data selection and backtranslation. 


We show how the different FDA data selection configurations tend to select different numbers of sentences coming from different systems, resulting in MT systems with different performance. 
 
Under the assumption that FDA's performance is hindered by the fact that the data originates from MT systems, and as such contains errors and is of lower lexical richness, we rescored the data selection scores for each sentence by a factor depending on the BLEU, TER and MTLD values of the system used to backtranslate it. By doing so, we managed to improve the results for the DE-EN system, while for EU-ES we obtained similar performance to the other MT systems; this allows us to use just 25\% of the data. Further investigation is required to study under which conditions our proposed rescoring method is beneficial, but our experiments with both low- and high-resource language pairs suggest that if the systems used for backtranslation are poor, then this technique will be of little value; clearly this is closely related to the amount of resources available for the language pair under study.


In the future, we plan to investigate ways to directly incorporate the rescoring metrics into the data selection process itself, so that penalising similar sentences can also be taken into account. We also aim to conduct a human evaluation of the translated sentences in order to obtain a better understanding of the effects of data selection and backtranslation on the overall quality. Finally, we intend to analyse the effect of these measures in a wider range of language pairs and settings, in order to propose a more general solution.

\section*{Acknowledgements}
Xabier Soto's work was supported by the Spanish Ministry of Economy and Competitiveness (MINECO) FPI grant number BES-2017-081045. This work was mostly done during an internship at the ADAPT Centre in DCU.

The ADAPT Centre for Digital Content Technology is funded under the Science Foundation Ireland (SFI) Research Centres Programme (Grant No. 13/RC/2106) and is co-funded under the European Regional Development Fund.

\bibliographystyle{acl_natbib}
\bibliography{acl2020}

\end{document}